\documentclass[11pt,a4paper]{article}
\usepackage[hyperref]{naaclhlt2019}

\usepackage{times}
\usepackage{latexsym}
\usepackage{url}
\usepackage{graphicx}
\graphicspath{{./images/}}
\usepackage{amsmath}
\usepackage{booktabs}
\usepackage{colortbl}
\usepackage{pifont}
\usepackage{subcaption}
\usepackage{tikz}
\usepackage{flushend}

\aclfinalcopy 
\def\aclpaperid{***} 

\newcommand{\utterance}[1]{\textit{``#1''}}
\newcommand{\phrase}[1]{\textit{`#1'}}
\newcommand{\struct}[1]{\texttt{\small #1}}

\newcommand{\squishlist}{
 \begin{list}{$\bullet$}
  { \setlength{\itemsep}{0pt}
     \setlength{\parsep}{3pt}
     \setlength{\topsep}{3pt}
     \setlength{\partopsep}{0pt}
     \setlength{\leftmargin}{1.0em}
     \setlength{\labelwidth}{1em}
     \setlength{\labelsep}{0.5em} } }

\newcommand{\squishend}{
  \end{list}  }

\title{ComQA: A Community-sourced Dataset for\\Complex Factoid Question Answering
with Paraphrase Clusters}

\author{Abdalghani Abujabal\textsuperscript{1}, Rishiraj Saha Roy\textsuperscript{2}, Mohamed Yahya\textsuperscript{3} and Gerhard Weikum\textsuperscript{2} \\
   \textsuperscript{1}Amazon Alexa, Aachen, Germany\\
{\tt abujabaa@amazon.de } \\
   \textsuperscript{2}Max Planck Institute for Informatics, Saarland Informatics Campus, Germany \\
     {\tt \{rishiraj, weikum\}@mpi-inf.mpg.de} \\
   \textsuperscript{3}Bloomberg L.P., London, United Kingdom\\
{\tt yahya.mohamed@gmail.com }
  \\}
\date{}

\begin{document}
\maketitle

\begin{abstract}
To bridge the gap between the capabilities of the state-of-the-art in factoid 
question answering (QA) and what users ask, we need large datasets of real  
user questions that capture the various question phenomena users are interested in,
and the diverse ways in which these questions are formulated.
We introduce \emph{ComQA}, a \emph{large} dataset of \emph{real} user 
questions that exhibit different \emph{challenging aspects} such as  \emph{compositionality}, 
\emph{temporal reasoning}, and \emph{comparisons}.
ComQA questions come from the WikiAnswers community QA platform, which typically contains questions
that are not satisfactorily answerable by existing search engine technology.
Through a large crowdsourcing effort, we clean the question dataset, group 
questions into \emph{paraphrase clusters}, and annotate clusters with their answers.
ComQA contains $11,214$ questions grouped into 4,834 paraphrase clusters.
We detail the process of constructing ComQA, including the measures taken to ensure its high quality while making effective use of crowdsourcing.
We also present an extensive analysis of the dataset 
and the results achieved by state-of-the-art systems on ComQA, demonstrating that our dataset can be a driver of future research on QA.
\end{abstract}
\footnotetext[1]{The main part of this work was carried out when the author
was at the Max Planck Institute for Informatics.}

\section{Introduction}
\label{sec:intro}

\begin{figure} [t]
  \centering
    \includegraphics[width=\columnwidth]{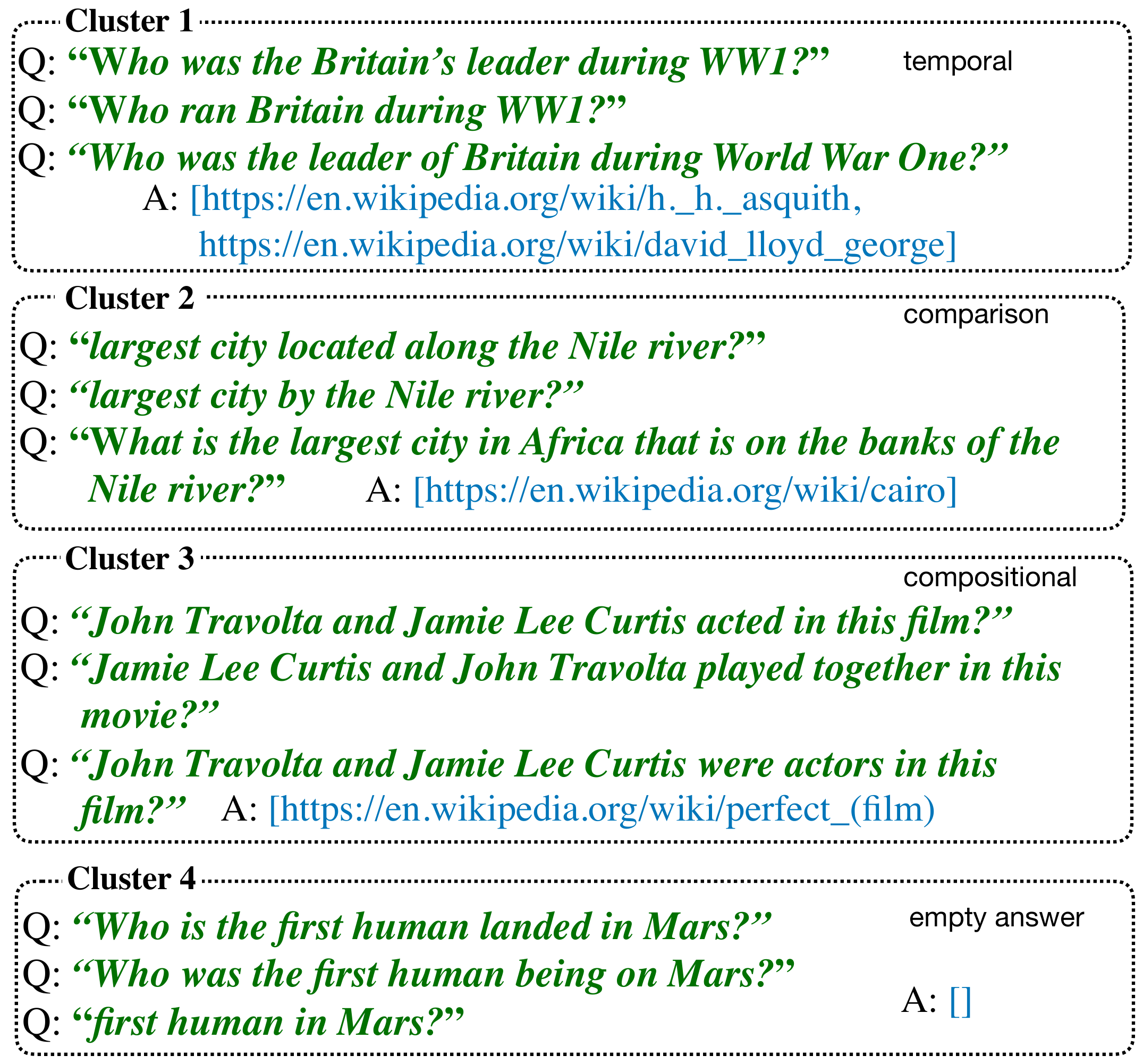}
      \caption{ComQA paraphrase clusters covering a range of question aspects e.g., temporal and compositional questions, with lexical and syntactic diversity.}
      \label{fig:exam}
\end{figure}

Factoid QA is the task of answering natural language questions whose answer is one or a small
number of entities \cite{DBLP:conf/sigir/VoorheesT00}.
To advance research
in QA in a manner consistent with the needs of end users, it is important to 
have access to datasets that reflect \textcolor{red}{\emph{real}} user information needs
by covering various \textcolor{brown}{\emph{question phenomena}} and the wide 
\textcolor{teal}{\emph{lexical and syntactic variety}} in expressing these information needs. The benchmarks should 
be \textcolor{violet}{large} enough to facilitate the use of data-hungry machine learning 
methods.
In this paper, we present ComQA, a \textcolor{violet}{large dataset of
11,214} \textcolor{red}{real} user questions collected from the WikiAnswers community QA website. 
As shown in Figure \ref{fig:exam}, the dataset contains \textcolor{brown}{various question phenomena}. 
ComQA questions are grouped into \textcolor{teal}{4,834 paraphrase clusters}
through a large-scale crowdsourcing effort, which capture lexical and syntactic variety.
Crowdsourcing is also used to pair 
paraphrase clusters with answers to serve as a supervision signal for 
training and as a basis for evaluation.

\begin{table*} [t]
  \centering
    \resizebox{\textwidth}{!}{%
    \begin{tabular}{l | c | c | c | c} \toprule
            \textbf{Dataset}  &       \textbf{\textcolor{violet}{Large scale ($>5$K)}}      & \textbf{\textcolor{red}{Real Information Needs}} & \textbf{ \textcolor{brown}{Complex Questions}} & \textbf{\textcolor{teal}{Question Paraphrases}}  \\ \midrule 
    \textbf{ComQA} (This paper) & \ding{51} & \ding{51} & \ding{51} & \ding{51}  \\ \bottomrule 
       Free917~\cite{cai:13}        & \ding{55}         & \ding{55} &   \ding{55} &   \ding{55}  \\ 
       WebQuestions~\cite{berant:13}       & \ding{51}         & \ding{51} & \ding{55}& \ding{55} \\ 
    SimpleQuestions~\cite{bordes:15}        & \ding{51}         & \ding{55} & \ding{55} & \ding{55} \\ 
    QALD~\cite{usbeck:17}        & \ding{55}       &  \ding{55} & \ding{51} & \ding{55} \\
    LC-QuAD~\cite{DBLP:conf/semweb/TrivediMDL17}        & \ding{51}         & \ding{55} & \ding{51} &  \ding{55} \\ 
    ComplexQuestions~\cite{DBLP:conf/coling/BaoDYZZ16} & \ding{55} & \ding{51}  & \ding{51} & \ding{55} \\ 
    GraphQuestions~\cite{DBLP:conf/emnlp/SuSSSGYY16} & \ding{51} & \ding{55}  & \ding{51} & \ding{51} \\ 
    ComplexWebQuestions~\cite{DBLP:journals/corr/abs-1803-06643} & \ding{51} & \ding{55} & \ding{51} & \ding{55} \\ 
    TREC~\cite{DBLP:conf/sigir/VoorheesT00}    & \ding{55} & \ding{51} & \ding{51} & \ding{55} \\ 
    \bottomrule

    \end{tabular} 
    }
  \caption{Comparison of ComQA with existing QA datasets over various dimensions.}
  \label{tab:comp}
\end{table*} 

Table~\ref{tab:comp} contrasts ComQA with publicly available QA
datasets. The foremost issue that ComQA tackles is ensuring research is driven by information needs formulated by real users. Most large-scale datasets resort to highly-templatic 
synthetically generated
natural language questions~\cite{bordes:15,cai:13,DBLP:conf/emnlp/SuSSSGYY16,DBLP:journals/corr/abs-1803-06643,DBLP:conf/semweb/TrivediMDL17}. 
Other datasets utilize search engine logs to collect their questions~\cite{berant:13}, 
which creates a bias towards simpler questions that search engines can already answer
reasonably well. 
In contrast, ComQA questions come from WikiAnswers, a community QA website where
users pose questions to be answered by other users. This is often a reflection
of the fact that such questions are beyond the capabilities of commercial search 
engines and QA systems.
Questions in our dataset exhibit a wide range of interesting aspects such as
the need for temporal reasoning (Figure \ref{fig:exam}, cluster 1),  comparison (Figure \ref{fig:exam}, cluster 2), compositionality (multiple subquestions with multiple entities and relations) (Figure \ref{fig:exam}, cluster 3), and unanswerable questions (Figure \ref{fig:exam}, cluster 4).

ComQA is the result of a carefully designed large-scale 
crowdsourcing effort to group questions into paraphrase clusters
and pair them with answers.
Past work has demonstrated the benefits of paraphrasing for QA~\cite{DBLP:conf/www/AbujabalRYW18,DBLP:conf/acl/BerantL14,DBLP:conf/emnlp/DongMRL17,DBLP:conf/acl/FaderZE13}.
Motivated by this, we judiciously use crowdsourcing to obtain clean paraphrase clusters from 
WikiAnswers' noisy ones, resulting in ones like those shown in Figure~\ref{fig:exam},
with both lexical and syntactic variations. The only other dataset to provide such 
clusters is that of Su et al. \shortcite{DBLP:conf/emnlp/SuSSSGYY16}, but that is based
on synthetic information needs.

For answering, recent research has shown that combining various resources for 
answering significantly improves performance~\cite{savenkov:16,sun2018open,
DBLP:conf/acl/XuRFHZ16}. Therefore, we do not pair ComQA with a specific knowledge base (KB) or text corpus for answering.
We call on the research community to innovate in combining different answering sources to 
tackle ComQA and advance research in QA.
We use crowdsourcing to pair paraphrase clusters with answers. ComQA
answers are primarily Wikipedia entity URLs. This has two motivations: (i)
it builds on the example of search engines that use Wikipedia entities as answers for entity-centric queries (e.g., through knowledge cards), and
(ii) most modern KBs ground their entities in Wikipedia. 
Wherever the answers are temporal or measurable quantities, we use TIMEX3\footnote{{\small \url{http://www.timeml.org}}} 
and the 
International System of Units\footnote{{\small\url{https://en.wikipedia.org/wiki/SI}}}
for normalization.
Providing canonical answers allows for better comparison of different systems.

We present an extensive analysis of ComQA, where we introduce
the various question aspects of the dataset. We also analyze the
results of running state-of-the-art QA systems on ComQA. 
ComQA exposes major shortcomings in these systems, mainly
related to their inability to handle compositionality, time, and comparison.
Our detailed error analysis provides inspiration for avenues of future work to
ensure that QA systems meet the expectations of real users.
To summarize, in this paper we make the following contributions:
\begin{itemize}

\item We present a dataset of 11,214 real user 
questions collected from a community QA website. 
The questions exhibit a range of
aspects that are important for users and challenging for existing QA systems.  
Using crowdsourcing, questions are grouped into 4,834 paraphrase clusters that 
are annotated with answers. 
ComQA is available at: {\small \url{http://qa.mpi-inf.mpg.de/comqa}}.

\item We present an extensive analysis and quantify the various
difficulties in ComQA. We also present the results of state-of-the art QA systems on ComQA, and a detailed error analysis. 

\end{itemize}

\section{Related Work}
\label{sec:related}

There are two main variants of the factoid QA task, with  
the distinction tied to the underlying answering resources and the nature of answers. Traditionally, QA has been explored over
large textual corpora~\cite{DBLP:conf/sigir/CuiSLKC05,DBLP:conf/acl/HarabagiuMPMSBGRM01,DBLP:journals/nle/HarabagiuMP03,DBLP:conf/acl/RavichandranH02,DBLP:journals/jair/SaqueteGMML09} with answers being textual phrases.
Recently, it has been explored over large structured resources 
such as KBs~\cite{berant:13,DBLP:conf/www/UngerBLNGC12}, with answers being semantic entities.
Recent work demonstrated that the two variants are complementary, and a combination of the two results in the best performance~\cite{sun2018open,DBLP:conf/acl/XuRFHZ16}. 

{\bf QA over textual corpora.}
QA has a long tradition in IR and NLP,
including benchmarking tasks in TREC~\cite{DBLP:conf/sigir/VoorheesT00,
dietz2017trec} and CLEF~\cite{DBLP:conf/clef/MagniniVAEPRRSS05,
DBLP:conf/clef/HerreraPV05}.
This has predominantly focused on retrieving answers
from textual sources~\cite{ferrucci:12,DBLP:conf/sigir/HarabagiuLH06,
DBLP:conf/acl/PragerCC04,
DBLP:conf/acl/SaqueteMMG04,yin:15}.
In IBM Watson~\cite{ferrucci:12}, structured data played
a role, but text was the main source for answers.
The TREC QA evaluation series provide hundreds of questions to be answered over documents, which have become widely adopted benchmarks for answer sentence selection
~\cite{DBLP:conf/acl/WangN15}.
ComQA is orders of magnitude larger than TREC QA. 

\emph{Reading comprehension} (RC) is a recently introduced task,
where the goal is to answer a question from a \emph{given} textual 
paragraph~\cite{DBLP:journals/corr/abs-1712-07040,DBLP:conf/emnlp/LaiXLYH17,DBLP:conf/emnlp/RajpurkarZLL16,DBLP:conf/rep4nlp/TrischlerWYHSBS17,DBLP:conf/emnlp/YangYM15}.
This setting is different from factoid QA, where the goal is to answer 
questions from a large repository of data (be it textual or structured), and
not a single paragraph. A recent direction 
in RC is dealing with unanswerable questions from the underlying data \cite{DBLP:conf/acl/RajpurkarJL18}. ComQA includes such questions to allow tackling the same problem in the context of factoid QA.

{\bf QA over KBs.}
Recent efforts have focused on natural language questions
as an interface for KBs,
where questions are translated to structured queries via semantic parsing
~\cite{DBLP:conf/coling/BaoDYZZ16,DBLP:conf/cikm/BastH15,
DBLP:conf/acl/FaderZE13,DBLP:conf/naacl/MohammedSL18,reddy:14,DBLP:conf/emnlp/YangDZR14,DBLP:conf/acl/YaoD14,DBLP:conf/cikm/YahyaBEW13}.
Over the past five years, many datasets were introduced for this setting. 
However, as Table~\ref{tab:comp} shows, they are either small in size 
(Free917, and ComplexQuestions), composed of synthetically generated 
questions (SimpleQuestions, GraphQuestions, LC-QuAD and ComplexWebQuestions), 
or are structurally simple (WebQuestions). 
ComQA
addresses
these shortcomings. 
Returning semantic entities as answers allows users
to further explore these entities in various resources such as their Wikipedia 
pages, Freebase entries, etc. It also allows
QA systems to tap into various interlinked resources for improvement (e.g.,
to obtain better lexicons, or train better NER systems). 
Because of this,
ComQA provides semantically grounded reference answers in Wikipedia (without committing to Wikipedia as an answering resource). 
For
numerical quantities and dates, ComQA adopts the International System of Units and TIMEX3 standards, respectively.
\section{Overview}
\label{sec:setup}

In this work, a factoid question is
a question whose answer is one or 
a small number of entities or literal values~\cite{DBLP:conf/sigir/VoorheesT00} e.g., \utterance{Who were the secretaries of state under 
Barack Obama?} and \utterance{When was Germany's first post-war chancellor born?}.

\subsection{Questions in ComQA}
\label{sec:ques}

A question in our dataset can exhibit
\emph{one or more} of the following phenomena:
\squishlist

\item \textbf{Simple:} questions  about a single property of
an entity (e.g., \utterance{Where was Einstein born?})

\item \textbf{Compositional:} A question is compositional if answering it
requires answering more primitive questions and combining these.
These can be \emph{intersection} or \emph{nested} questions. 
Intersection questions are ones where two or more subquestions
can be answered independently, and their answers intersected (e.g., 
\utterance{Which films featuring Tom Hanks did Spielberg direct?}).
In nested questions, the answer to one subquestion is necessary to 
answer another (\utterance{Who were the parents of the thirteenth president of the US?}).

\item \textbf{Temporal:} These are questions that require temporal reasoning for
deriving the answer, be it explicit (e.g., \phrase{in 1998}), implicit 
(e.g., \phrase{during the WWI}), relative (e.g., \phrase{current}), or
latent (e.g. \phrase{Who is the US president?}). Temporal questions also include
those \textit{whose answer} is an explicit temporal expression
(\utterance{When did Trenton become New Jersey's capital?}).

\item \textbf{Comparison:} We consider three types of comparison questions: \emph{comparatives}
(\utterance{Which rivers in Europe are longer than the Rhine?}),
\emph{superlatives} (\utterance{What is the population of the largest city
in Egypt?}), and \emph{ordinal} questions (\utterance{What was the name
of Elvis's first movie?}).

\item \textbf{Telegraphic}~\cite{DBLP:conf/emnlp/JoshiSC14}: These are
short questions formulated in an informal
manner similar to keyword queries (\utterance{First president India?}).
Systems that rely on linguistic analysis
often fail on such questions.

\item \textbf{Answer tuple}: Where an answer is a tuple of connected entities
as opposed 
to a single entity (\utterance{When and where did George H. Bush
go to college, 
and what did he study?}).
\squishend

\subsection{Answers in ComQA}
\label{sec:ans}
Recent work has shown that the choice of answering resource, or the combination of resources significantly
affects answering performance~\cite{savenkov:16,sun2018open,DBLP:conf/acl/XuRFHZ16}. 
Inspired by this, ComQA is not tied to a specific resource for answering. 
To this end, answers in ComQA are primarily Wikipedia URLs. 
This enables QA systems to combine different answering resources which
are linked to Wikipedia (e.g., DBpedia, Freebase, YAGO, Wikidata, etc). This also allows seamless comparison across these QA systems.
An answer in ComQA can be:
\squishlist

\item \textbf{Entity:} ComQA entities are  grounded in Wikipedia. However, Wikipedia
is inevitably incomplete, so answers that cannot be grounded
in Wikipedia are represented as plain text. For example, the answer for \utterance{What is the name of Kristen Stewart adopted brother?}
is \{Taylor Stewart, Dana Stewart\}.

\item \textbf{Literal value:} 
Temporal answers follow the TIMEX3 standard.
For measurable quantities, we follow the International System of Units. 

\item \textbf{Empty:} In the factoid setting, some questions can be based on a false premise, and hence, are unanswerable e.g.,
~\utterance{Who was the first human being on Mars?} (no human has been on Mars, yet). The correct answer to such questions is the empty set. 
Such questions allow systems to cope with these cases. 
Recent work has started looking at this problem~\cite{DBLP:conf/acl/RajpurkarJL18}.
\squishend
\section{Dataset Construction}
\label{sec:collection}

Our goal is to collect factoid questions that represent real information needs and cover a range of question aspects. Moreover, 
we want to have different paraphrases for each question. 
To this end, we tap into the potential of community QA platforms.
Questions posed there represent real information needs.
Moreover, users of those platforms provide (noisy)
annotations around questions e.g., paraphrase clusters.
In this work, we exploit the annotations where users mark questions
as duplicates as a basis for paraphrase clusters, and clean those.
Concretely, we started with the WikiAnswers crawl by Fader
et al.~\shortcite{DBLP:conf/kdd/FaderZE14}. We obtained ComQA from this crawl
primarily through a large-scale crowdsourcing effort, which we describe in what follows.

The original resource curated by Fader et al. contains $763$M questions.
Questions in the crawl are grouped into $30$M
paraphrase clusters based on feedback from WikiAnswers users.
This clustering has
a low accuracy~\cite{DBLP:conf/kdd/FaderZE14}.
Extracting factoid questions and cleaning the clusters are thus essential
for a high-quality dataset. 

\subsection{Preprocessing of WikiAnswers}
\label{subsec:preprocess}

To remove non-factoid questions, we filtered out questions that (i) start with \phrase{why}, or (ii) contain words like \textit{(dis)similarities, differences, (dis)advantages}, etc. Questions matching these filters are out of scope as they require a narrative answer. 
We also removed questions with less than three or more than twenty words, as we found these to be typically noisy or non-factoid questions. 
This left us with about $21$M questions belonging to $6.1$M clusters.

To further focus on factoid questions, we automatically classified questions into one or more of the following \emph{four classes}: 
(1) temporal, (2) comparison, (3) single entity, and (4) multi-entity questions. We used
SUTime~\cite{DBLP:conf/lrec/ChangM12} to identify temporal questions
and the Stanford named entity recognizer~\cite{DBLP:conf/acl/FinkelGM05}
to detect named entities.
We used part-of-speech patterns to identify comparatives, superlatives, and 
ordinals.
Clusters which did not have questions belonging to any of the 
above classes were discarded from further consideration.
Although these clusters contain false negatives e.g., \utterance{What official position did Mendeleev hold until his death?}
due to errors by the tagging tools, most discarded questions are out-of-scope.

\textbf{Manual inspection.} 
We next applied the first stage of human curation to the dataset. 
Each WikiAnswers cluster was assigned to one of the four classes above based on the majority label of the questions within. 
We then randomly sampled $15$K clusters
from each of the four classes ($60$K clusters in total with $482$K questions) and 
sampled a representative question from each of these clusters at random
($60$K questions).
We relied on the assumption that questions within the same cluster
are semantically equivalent.
These $60$K questions were
manually {examined by the authors} and those with unclear or non-factoid
intent were removed along with the cluster that contains them.
We thus ended up with $2.1$K clusters with $13.7$K questions.

\subsection{Curating Paraphrase Clusters}
\label{subsec:taks1}

\begin{figure} [t]
  \centering
    \includegraphics[width=0.9\columnwidth]{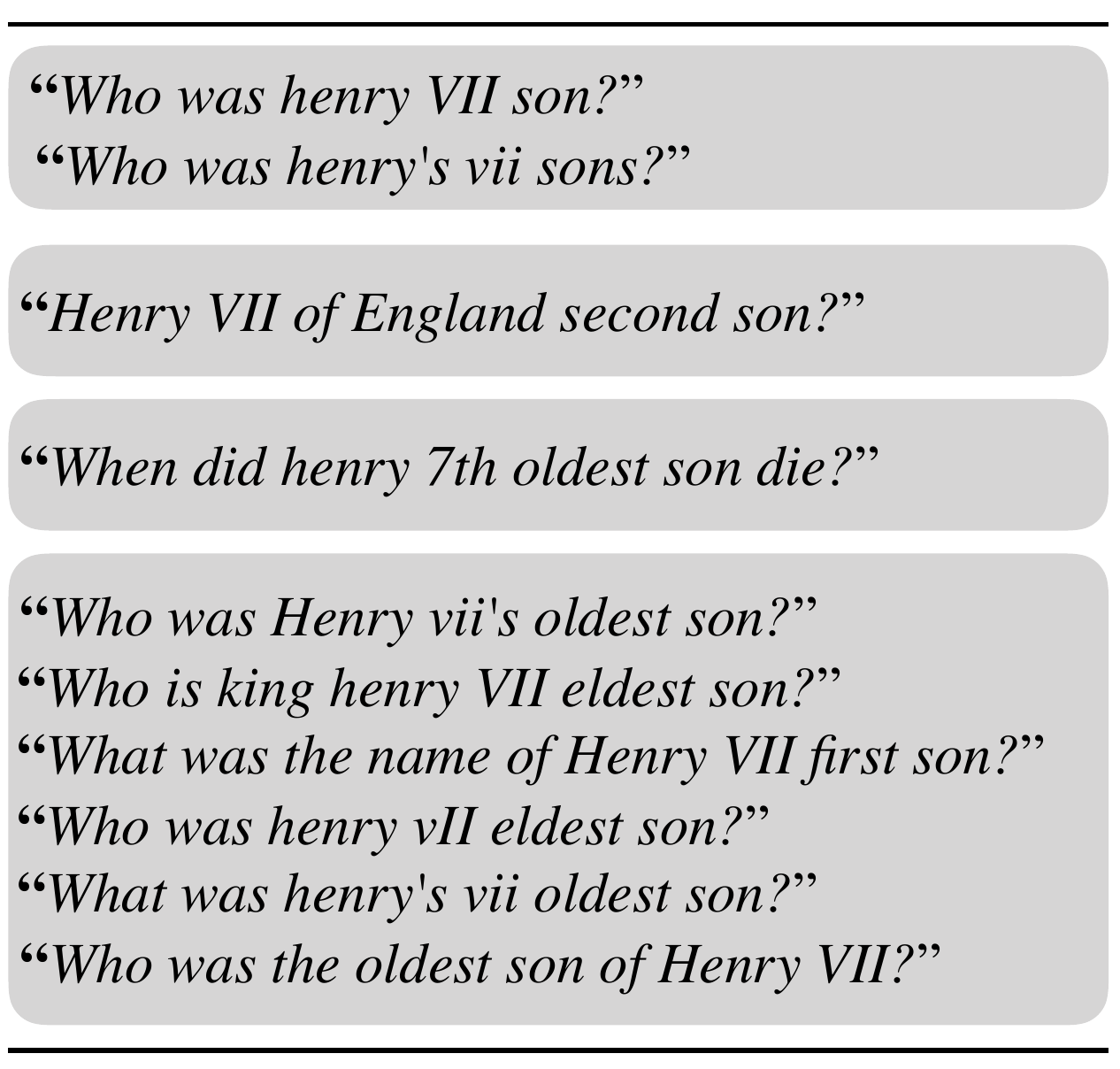}
      \caption{A WikiAnswers cluster split into four clusters by AMT Turkers.}
      \label{fig:clus}
\end{figure}

We inspected a random subset of the $2.1$K WikiAnswers clusters and found
that questions in the same cluster are semantically \textit{related
but not necessarily equivalent}, which is
in
line with observations in previous work
~\cite{DBLP:conf/kdd/FaderZE14}. Dong et al. \shortcite{DBLP:conf/emnlp/DongMRL17} reported that 45\% of question pairs
were related rather than genuine paraphrases.
For example, Figure~\ref{fig:clus} shows 10 questions in the same WikiAnswers cluster.
Obtaining accurate paraphrase clusters is crucial to any systems that want to 
utilize them~\cite{DBLP:conf/www/AbujabalRYW18,
DBLP:conf/acl/BerantL14,
DBLP:conf/emnlp/DongMRL17}.
We therefore utilized crowdsourcing to clean the Wikianswers
paraphrase clusters. 
We used Amazon Mechanical Turk (AMT) to identify semantically equivalent 
questions
within a WikiAnswers cluster, thereby obtaining cleaner clusters for ComQA.
Once we had the clean clusters, we set up a second AMT task to collect answers
for each ComQA cluster.

\textbf{Task design.} 
We had to ensure the simplicity of the task to obtain high quality results.
Therefore, rather than giving workers a WikiAnswers cluster and asking them 
to partition it into clusters of paraphrases, we showed them pairs of 
questions from a cluster and asked them to make the binary decision of whether the two questions are
paraphrases.
To reduce potentially redundant annotations, we utilized
the transitivity
of the paraphrase relationship. Given a WikiAnswers cluster
$Q=\{q_1, ..., q_n\}$,
we proceed in rounds to form ComQA clusters. In the first round, 
we collect annotations for each pair $(q_i, q_{i+1})$. 
The majority annotation among five annotators is taken. 
An initial clustering is formed accordingly, with clusters sharing the same
question merged together (to account for transitivity).
This process continues iteratively until no new clusters can be formed from $Q$.

\begin{table*} [t!] \small 
  \centering
	  \resizebox{\textwidth}{!}{%
		  \begin{tabular}{l l r} \toprule
          \textbf{Property}  						& \textbf{Example}																							& \textbf{Percentage\%}	\\ \toprule
	      \textit{Compositional questions} 			& 																											& 						\\ \midrule  
	      Conjunction 								& \utterance{What is the capital of the country whose northern border is Poland and Germany?}				& $17.67$ 				\\   
	      Nested 									& \utterance{When is Will Smith's oldest son's birthday?}   												& $14.33$   			\\ \midrule
	      \textit{Temporal questions} 				& 																											& 						\\ \midrule
	      Explicit time        						& \utterance{Who was the winner of the World Series \textbf{ in 1994?}}      								& $4.00$  				\\ 
	      Implicit time        						& \utterance{Who was Britain's leader \textbf{during WW1?}}      											& $4.00$  				\\ 
	      Temporal answer 							& \utterance{\textbf{When} did Trenton become New Jersey's capital?}    									& $15.67$   			\\ \midrule
	      \textit{Comparison questions} 			& 																											& 						\\ \midrule
	      Comparative 								& \utterance{Who was the first US president to serve \textbf{2 terms}?}     								& $1.00$  				\\ 
	      Superlative 								& \utterance{What ocean does the \textbf{longest} river in the world flow into?}    						& $14.33$   			\\ 
	      Ordinal 									& \utterance{When was Thomas Edisons \textbf{first} wife born?}    											& $14.00$   			\\ \midrule  
		  \textit{Question formulation} 			& 																											& 						\\ \midrule
		  Telegraphic 								& \utterance{Neyo first album?}    																			& $8.00$   				\\ \midrule  
	      \textit{Entity distribution in questions} & 																											& 						\\ \midrule
	      Zero entity        						& \utterance{What public company has the most employees in the world?}    									& $2.67$   				\\ 
          Single entity        						& \utterance{Who is \textbf{Brad Walst}'s wife?}     														& $75.67$  				\\ 
		  Multi-entity        						& \utterance{What country in \textbf{South America} lies between \textbf{Brazil} and \textbf{Argentina}?}	& $21.67$ 				\\ \midrule 
	      \textit{Other features} 					& 																											& 						\\ \midrule 
	      Answer tuple 								& \utterance{\textbf{Where} was Peyton Manning born and \textbf{what year was he born}?}   					& $2.00$  				\\ 
	      Empty answer 								& \utterance{Who was Calgary's first woman mayor?} 														& $3.67$ 				\\ \bottomrule
    \end{tabular} }
  \caption{Results of the manual analysis of $300$ questions. Note that
  properties are not mutually exclusive.}
  \label{tab:examples}
\end{table*}

\begin{figure} [t]
  \centering
    \includegraphics[width=1\columnwidth]{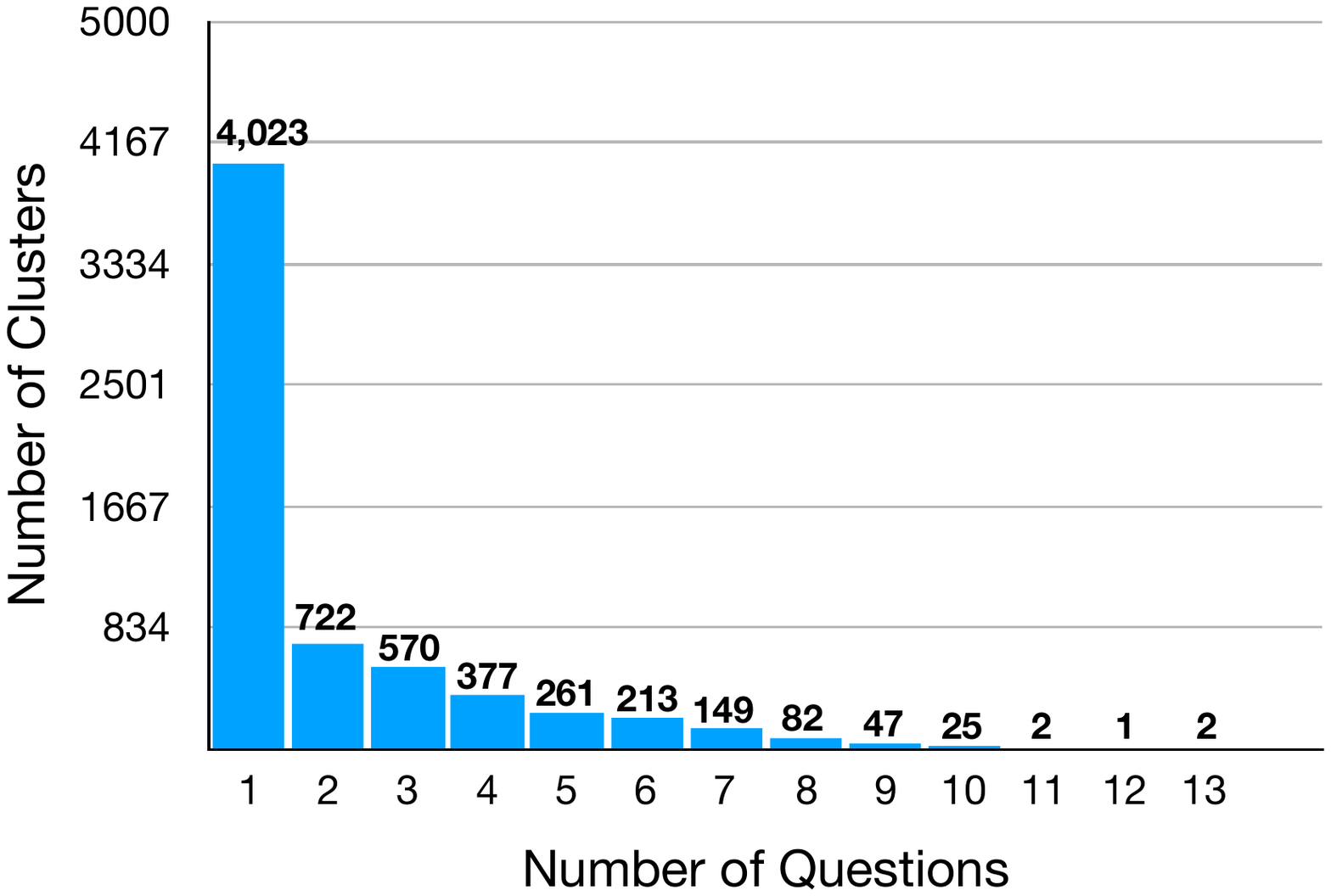}
      \caption{The distribution of questions in clusters.}
      \label{fig:hist}
\end{figure}

\textbf{Task statistics.} We obtained annotations for 18,890 question
pairs from 175 different workers. Each pair was shown to five different workers, with
$65.7\%$ of the pairs receiving unanimous agreement, $21.4\%$ receiving
four agreements and $12.9\%$ receiving three agreements.
By design, 
with five judges and binary annotations, no pair can have less three agreements.
This resulted in questions being placed in paraphrase clusters,
and no questions were discarded at this stage.
At the end of this step, the original $2.1$K WikiAnswers
clusters became $6.4$K ComQA clusters  with a total of $13.7$K questions.
Figure~\ref{fig:hist} shows the distribution of questions in clusters.

To test whether relying on the transitivity of the paraphrase 
relationship is suitable to reduce the annotation effort, we asked
annotators to annotate 1,100 random pairs $(q_1, q_3)$, where we had already 
received positive annotations for the pairs $(q_1, q_2)$ and $(q_2, q_3)$
being paraphrases of each other. In $93.5$\% of the cases there was 
agreement. 
Additionally, as experts on the task, the authors manually assessed $600$ pairs
of questions, which serve as honeypots. 
There was $96.6\%$ agreement with our annotations.
An example result of this task is shown in Figure~\ref{fig:clus}, where
Turkers split the original WikiAnswers cluster into the four clusters shown.

\subsection{Answering Questions}
\label{subsec:task2}

We were now in a position to obtain an answer
annotation for each of the $6.4$K clean clusters.

\textbf{Task design.} 
To collect answers, we designed another AMT task, where workers were shown a representative question randomly drawn from a cluster. Workers were asked to use the Web to find answers and to provide the corresponding URLs of Wikipedia
entities. 
Due to the inevitable incompleteness of Wikipedia, workers were asked to
provide the surface form of an answer entity if it does not have a 
Wikipedia page.
If the answer is a full date, workers were asked to follow dd-mmm-yyyy format. 
For measurable quantities, workers were asked to provide units.
We use TIMEX3 and the international system of units for normalizing
temporal answers and measurable quantities e.g.,~\phrase{12th century} to
\struct{11XX}.
If no answer is found, workers were
asked to type in \phrase{no answer}.

\textbf{Task statistics.} 
Each representative question was shown to three different workers.
An answer is deemed correct if it is common between at least two workers.
This resulted in $1.6$K clusters (containing $2.4$K questions) with no agreed-upon answers, which were dropped.
For example, \utterance{Who was the first democratically elected president of Mexico?} is subjective.
Other questions received related answers e.g., \utterance{Who do the people in Iraq worship?} with
\struct{Allah}, \struct{Islam} and \struct{Mohamed} as answers from the three annotators.
Other questions were underspecified e.g., \utterance{Who was elected the vice president in 1796?}. 
At the end of the task, we ended up with 4,834 clusters with 11,214 question-answer pairs,
which form ComQA.
\section{Dataset Analysis}
\label{sec:analysis}

In this section, we present a manual analysis of $300$ questions sampled 
at random from the ComQA dataset.
This analysis helps understand the different aspects of our dataset.
A summary of the analysis is presented in Table~\ref{tab:examples}.

{\bf Question categories.}
We categorized each question as either simple or complex.
A question is complex if it belongs to one or more of
the compositional, temporal, or comparison classes. 
$56.33\%$ of the questions were complex; $32\%$ compositional,
$23.67\%$ temporal, and $29.33\%$ contain comparison conditions.
A question may contain multiple conditions 
(\utterance{What country has the \textbf{highest} population in the year \textbf{2008}?} with comparison and temporal conditions).

We also identified questions of telegraphic nature e.g.,
\utterance{Julia Alvarez's parents?}, with
$8\%$ of our questions being telegraphic.
Such questions pose a challenge for systems that rely on linguistic analysis of
questions~\cite{DBLP:conf/emnlp/JoshiSC14}.

We counted the number of named entities in questions: $23.67\%$ contain
two or more entities, reflecting their compositional nature, and
$2.67\%$ contain no entities e.g., \utterance{What public company has 
the most employees in the world?}.
Such questions can be hard as many methods assume the existence of
a pivot entity in a question.

Finally, $3.67\%$ of the questions are unanswerable, e.g.,
\utterance{Who was the first human being on Mars?}. 
Such questions incentivise QA systems to return non-empty answers only when
suitable. 
In Table ~\ref{tab:comp-cat} we compare ComQA with other current datasets based on real user information needs over different 
question categories.

\begin{figure} [t]
    \centering
    \begin{subfigure}[b]{0.49\columnwidth}
        \includegraphics[width=\columnwidth]{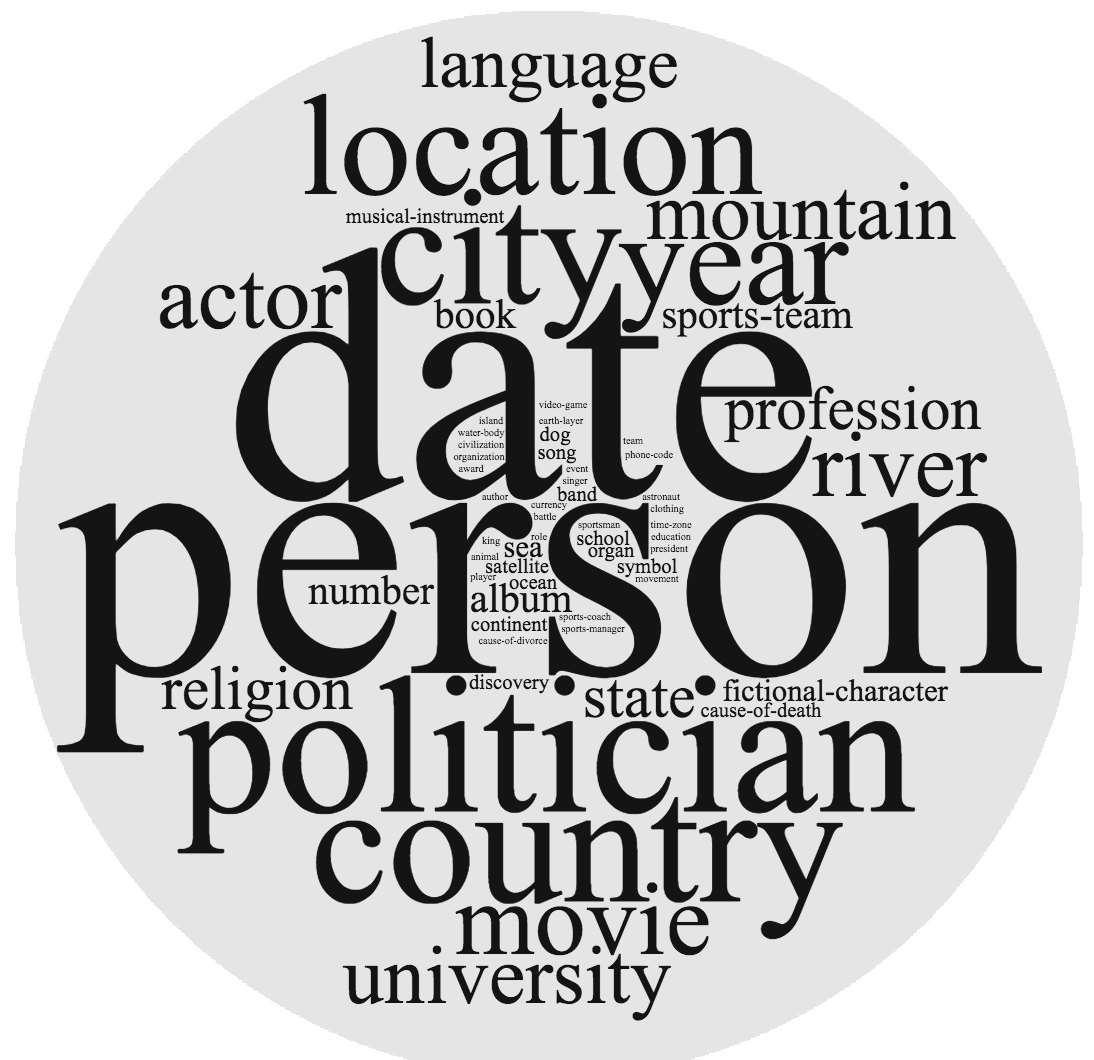}
        \caption{Answer types}
        \label{fig:wq-f1}
    \end{subfigure}
    \begin{subfigure}[b]{0.49\columnwidth}
        \includegraphics[width=\columnwidth]{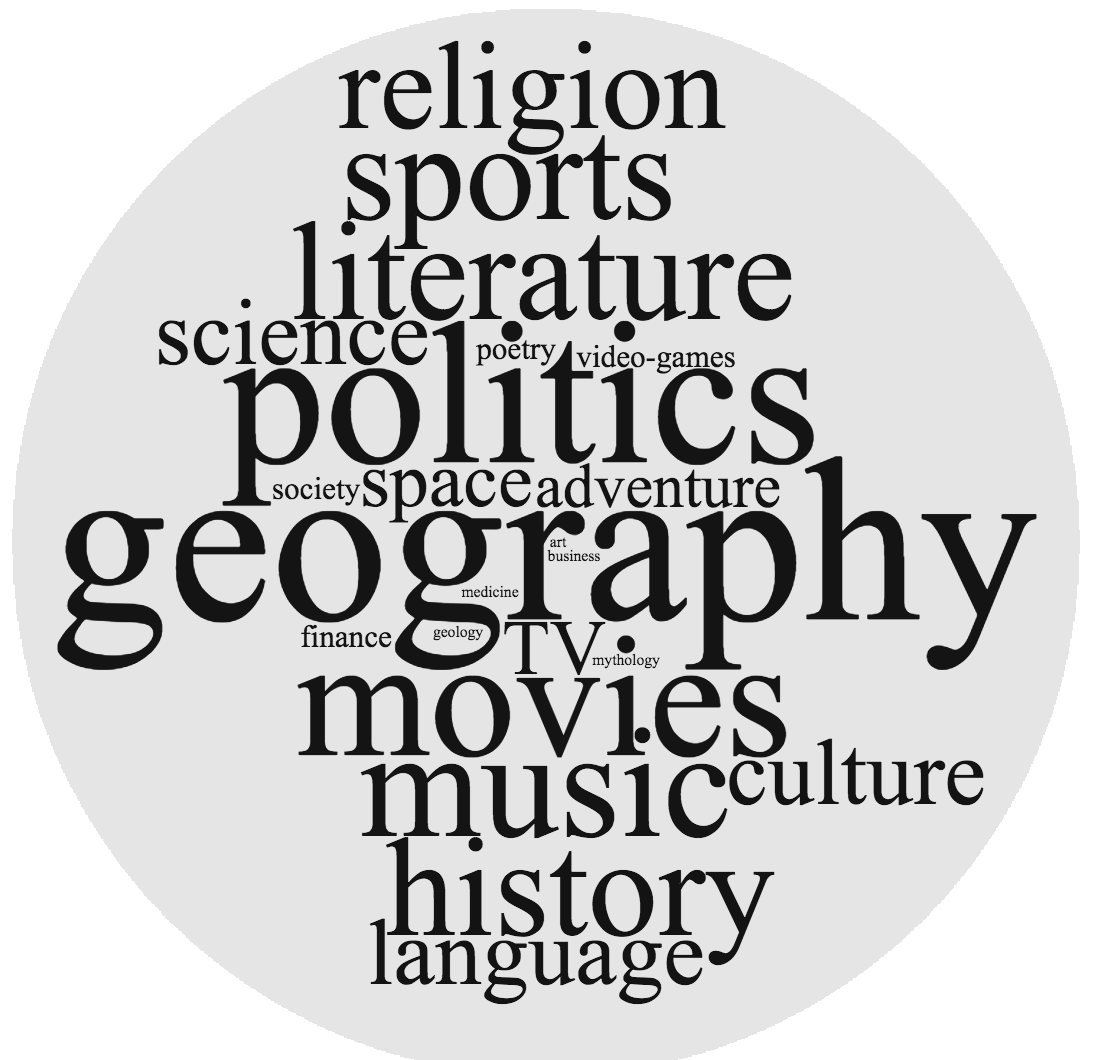}
        \caption{Question topics}
        \label{fig:wq-temps}
    \end{subfigure}
    \caption{Answer types and question topics on $300$ annotated examples 
    as word clouds.}
    \label{fig:type-topic-dist}
\end{figure}

{\bf Answer types.}
We annotated each question with the most fine-grained context-specific
answer type~\cite{ziegler:17}.
Answers in ComQA belong to a diverse set of types that
range from coarse (e.g., \struct{person})
to fine (e.g., \struct{sports manager}).
Types also include literals such as  \struct{number} and  \struct{date}.
Figure~\ref{fig:type-topic-dist}(a) shows answer types of the $300$ annotated examples as a word cloud.

{\bf Question topics.}
We annotated questions with topics to which they belong (e.g., geography, movies, sports).
These are shown in Figure~\ref{fig:type-topic-dist}(b), and demonstrate
the topical diversity of ComQA.

{\bf Question length.}
Questions in ComQA are fairly long, with a mean length of 7.73 words, indicating the compositional nature of questions.

\begin{table*} [t]  
	\centering
	    \resizebox{\textwidth}{!}{%
		\begin{tabular}{l | c | c | c | c | c | c} \toprule
       \textbf{Dataset}  									& \textbf{Size}  & \textbf{Compositional}   & \textbf{Temporal} & \textbf{Comparison}	& \textbf{Telegraphic}	& \textbf{Empty Answer}	\\ \toprule
	    \textbf{ComQA}  									& $11,214$ 		 & $32\%$ 					& $24\%$ 			& $30\%$  				& $8\%$ 				& $4\%$  				\\ \bottomrule
        WebQuestions~\cite{berant:13}      					& $5,810$ 	     & $2\%$ 				    & $7\%$ 			& $2\%$  				& $0\%$ 				& $0\%$  				\\ 
	    ComplexQuestions~\cite{DBLP:conf/coling/BaoDYZZ16}	& $2,100$ 		 & $39\%$ 					& $34\%$ 			& $9\%$ 				& $0\% $ 				& $0\% $  				\\ \bottomrule
  \end{tabular} }
  \caption{Comparison of ComQA with existing datasets over various phenomena. We manually annotated $100$ random questions from each dataset.}
  \label{tab:comp-cat}
\end{table*}

\section{Experiments}
\label{sec:experiments}

In this section we present experimental results 
for running ComQA through state-of-the-art QA systems. Our experiments
show that these systems achieve humble performance on ComQA. Through 
a detailed analysis, this performance can be attributed to 
systematic shortcomings in handling various question aspects in ComQA.

\subsection{Experimental Setup}
\label{subsec:setup}

{\bf Splits.}
We partition ComQA into a random train/dev/test split of 70/10/20\% with 7,850, 1,121 and 2,243 questions, respectively.

{\bf  Metrics.}
We follow the community's standard evaluation metrics:
we compute average precision, recall, and F1 scores across all test questions.
For unanswerable questions whose 
correct answer is the empty set,
we define precision and recall to be $1$ for a system that returns an 
empty set, and $0$ otherwise~\cite{DBLP:conf/acl/RajpurkarJL18}.

\subsection{Baselines}
\label{subsec:baselines}

We evaluated two categories of QA systems that differ in the underlying
answering resource:
either KBs or textual extractions.
We ran the following systems: 
(i) Abujabal et al.~\shortcite{DBLP:conf/www/AbujabalYRW17}, which automatically generates
templates using question-answer pairs;
(ii) Bast and Haussmann~\shortcite{DBLP:conf/cikm/BastH15}, which
instantiates hand-crafted query templates followed by query ranking;
(iii) Berant and Liang~\shortcite{berant2015agenda}, which relies on
agenda-based parsing and imitation learning; (iv) Berant et al.~\shortcite{berant:13}, which uses rules  to build
queries from questions; and
(v) Fader et al.~\shortcite{DBLP:conf/acl/FaderZE13}, which
maps questions to queries over open vocabulary facts extracted from Web documents. 
Note that our intention is not to assess the quality of these systems, but to assess how challenging ComQA is.

The systems were trained with ComQA data. All systems were run over the data sources for which they 
were designed.
The first four baselines are over Freebase. We therefore mapped 
ComQA answers (Wikipedia entities) to the corresponding Freebase names using
the information stored with entities in Freebase.
We observe that the Wikipedia 
answer entities have no counterpart in Freebase for  $7\%$ of the ComQA questions.
This suggests an oracle F1 score of $93.0$.
For Fader et al.~\shortcite{DBLP:conf/acl/FaderZE13}, which is over web extractions, we mapped Wikipedia URLs to their titles.

\begin{table} [t] 
  \centering
    \resizebox{\columnwidth}{!}{
    \begin{tabular}{l | c c c} \toprule
									                            & \textbf{Avg. Prec}	& \textbf{Avg. Rec} & \textbf{Avg. F1}  \\ \toprule
       Abujabal et al.~\shortcite{DBLP:conf/www/AbujabalYRW17}  & $21.2$         		& $38.4$    		& $22.4$    		\\
       Bast and Haussmann~\shortcite{DBLP:conf/cikm/BastH15}	&  $20.7$ 				& $37.6$ 			& $21.6$			\\        
       Berant and Liang~\shortcite{berant2015agenda}         	& $10.7$         		& $15.4$    		& $10.6$    		\\
       Berant et al.~\shortcite{berant:13}         				& $13.7$         		& $20.1$    		& $12.0$  			\\
       Fader et al.~\shortcite{DBLP:conf/acl/FaderZE13}    		&  $7.22$ 			& $6.59$ 			& $6.73$			\\ \bottomrule
    \end{tabular} }
  \caption{Results of baselines on ComQA test set.}
  \label{tab:baselines}
\end{table} 

\begin{table} [t] 
  \centering
   \resizebox{\columnwidth}{!}{
    \begin{tabular}{l | c c c} \toprule
													                & \textbf{WebQuestions}	& \textbf{Free917} 	& \textbf{ComQA}	\\ 
												                    & \textbf{F1}      		& \textbf{Accuracy} & \textbf{F1}  		\\ \toprule
	   Abujabal et al.~\shortcite{DBLP:conf/www/AbujabalYRW17}		& $51.0$         		& $78.6$    		& $22.4$  			\\ 
       Bast and Haussmann~\shortcite{DBLP:conf/cikm/BastH15}      	& $49.4$ 				& $76.4$ 			& $21.6$			\\        
       Berant and Liang~\shortcite{berant2015agenda}         		& $49.7$         		& $-$    			& $10.6$  			\\ 
       Berant et al.~\shortcite{berant:13}        					& $35.7$         		& $62.0$    		& $12.0$  			\\ \bottomrule
    \end{tabular}
   }
  \caption{Results of baselines on different datasets.}
  \label{tab:baselines-datasets}
\end{table} 

\subsection{Results}
\label{subsec:results}
Table~\ref{tab:baselines} shows the performance of the baselines on
the ComQA test set.
Overall, the systems achieved poor performance, suggesting that
current methods cannot handle the complexity of our dataset, and that
new models for
QA are needed.
Table~\ref{tab:baselines-datasets} compares the performance of the systems
on different datasets (Free917 uses accuracy as a quality metric).
For example, while
Abujabal et al.~\shortcite{DBLP:conf/www/AbujabalYRW17} achieved an F1 score
of $51.0$ on WebQuestions, it
achieved $22.4$ on ComQA.

The performance of Fader et al.~\shortcite{DBLP:conf/acl/FaderZE13} is worse
 than the others
due to the incompleteness of its underlying extractions and the complexity
of ComQA questions
that require higher-order relations and reasoning.
However, the system answered some complex questions, which KB-QA systems
failed to answer. For example, it answered \utterance{What is the highest mountain in the state of Washington?}. The answer to such a
question is more readily available in Web text, 
compared to a KB, where more sophisticated reasoning is required to handle the superlative.
However, a slightly modified question such as 
\utterance{What is the \textbf{fourth} highest mountain in the state of Washington?} is unlikely to be found in text, 
but be answered using KBs with the appropriate reasoning.
Both examples above demonstrate the benefits of combining text and structured resources.

\subsection{Error Analysis}
\label{subsec:error}
For the two best performing systems on ComQA, QUINT \cite{DBLP:conf/www/AbujabalYRW17} 
and AQQU \cite{DBLP:conf/cikm/BastH15}, we manually inspected 100 questions on which they failed.
We classified failure sources into four categories: compositionality, temporal, comparison or NER.
Table~\ref{tab:error} shows the distribution of these failure  sources.  

\vspace{3mm}
{\bf Compositionality. }
Neither system could handle the compositional nature of questions.
For example, they returned the father of Julius Caesar as an answer for \utterance{What did Julius Caesar's father work as?}, while, the question requires another KB predicate that connects the father to his profession.
For \utterance{John Travolta and Jamie Lee Curtis starred in this movie?}, 
both systems returned movies with  Jamie Lee Curtis, ignoring the constraint that John Travolta should also appear in them.
Properly answering multi-relation questions over KBs remains an open problem.

\vspace{2mm}
{\bf Temporal.}
Our analysis reveals that both systems fail to capture temporal constraints in questions, be it explicit or implicit.
For \utterance{Who won the Oscar for Best Actress in 1986?}, they returned all winners and ignored the temporal restriction from \phrase{in 1986}.
Implicit temporal constraints like named events (e.g., \phrase{Vietnam war} in \utterance{Who was the president of the US during Vietnam war?})
pose a challenge to current methods. Such constraints need to be detected 
first
and normalized to a canonical time interval (November 1st, 1955 to April 30th, 1975, for the Vietnam war).
Then, systems need to compare the terms of the US presidents with the above interval to 
account for the temporal relation of \phrase{during}.
While detecting explicit time expressions can be done reasonably well using existing time taggers~\cite{DBLP:conf/lrec/ChangM12}, identifying implicit ones is difficult.
Furthermore, retrieving the correct temporal scopes of  entities in questions (e.g., the terms of the US presidents) is hard due to the large number of temporal KB predicates associated with entities.

\vspace{2mm}
{\bf Comparison. }
Both systems perform poorly on comparison questions, which is expected since they were not designed to address those.
To the best of our knowledge, no existing KB-QA system can handle comparison questions.
Note that our goal is not to assess the quality the of current methods, but to highlight that these methods miss categories of questions that are important to real users.
For \utterance{What is the first film Julie Andrews made?} and \utterance{What is the largest city in the state of Washington?}, both systems
returned the list of  Julie Andrews's films and the list of Washington's cities, for the first and the second questions, respectively.
While the first question requires the attribute of \struct{filmReleasedIn} to order by, the second needs
the attribute of \struct{hasArea}. Identifying the correct attribute to order by as well as determining the order direction (ascending for the first and descending for the second) is challenging and out of scope for current methods.

\vspace{2mm}
{\bf NER. }
NER errors come from false negatives, where entities
are not detected. For example, in \utterance{On what date did the Mexican Revolution end?} QUINT identified \phrase{Mexican} rather than \phrase{Mexican Revolution} as an entity.
For \utterance{What is the first real movie that was produced in 1903?}, which does not ask about a specific entity, QUINT could not generate SPARQL queries. Existing QA methods expect a pivotal entity in a question, which is not always the case.

Note that while baseline systems achieved low
precision, they achieved higher recall (21.2 vs
38.4 for QUINT, respectively) (Table 4). This reflects the fact that these systems often cannot cope with the full complexity of ComQA questions, and instead end up evaluating underconstrained interpretations of the question.

To conclude, current methods can handle simple questions very well, but struggle with complex questions that involve
multiple conditions on different entities or need to join the results from sub-questions. Handling such complex questions, however, is 
important if we are to satisfy information needs expressed 
by real users.

\begin{table} [t] 
  \centering
   \resizebox{\columnwidth}{!}{
    \begin{tabular}{l | c | c} \toprule
		\textbf{Category} 	& \textbf{QUINT}		& \textbf{AQQU} \\  \toprule
	   Compositionality error		& $39\%$        & 		$43\%$ \\ 
       Missing comparison         		& $31\%$    &   	$26\%$ 	\\ 
       Missing temporal constraint       	& $19\%$		&	$22\%$ \\        
       NER error        					& $11\%$      & 	$9\%$ 	\\ \bottomrule
    \end{tabular}
   }
  \caption{Distribution of failure sources on ComQA questions on which QUINT and AQQU failed.}
  \label{tab:error}
\end{table} 
\section{Conclusion}
\label{sec:conclusion}

We presented ComQA, a dataset for QA that harnesses a community QA platform, reflecting questions asked by real users.
ComQA contains 11,214 question-answer pairs, with questions grouped into
paraphrase clusters through crowdsourcing.
Questions exhibit different aspects that current QA systems struggle with. 
ComQA is a challenging dataset that is aimed at driving future
research on QA, to match the needs of real users.

\section{Acknowledgment}
We would like to thank Tommaso Pasini for his helpful feedback.

\bibliographystyle{acl_natbib}
\bibliography{comqa}

\begin{thebibliography}{50}
\expandafter\ifx\csname natexlab\endcsname\relax\def\natexlab#1{#1}\fi

\bibitem[{Abujabal et~al.(2018)Abujabal, Roy, Yahya, and
  Weikum}]{DBLP:conf/www/AbujabalRYW18}
Abdalghani Abujabal, Rishiraj~Saha Roy, Mohamed Yahya, and Gerhard Weikum.
  2018.
\newblock Never-ending learning for open-domain question answering over
  knowledge bases.
\newblock In \emph{WWW}, pages 1053--1062.

\bibitem[{Abujabal et~al.(2017)Abujabal, Yahya, Riedewald, and
  Weikum}]{DBLP:conf/www/AbujabalYRW17}
Abdalghani Abujabal, Mohamed Yahya, Mirek Riedewald, and Gerhard Weikum. 2017.
\newblock Automated template generation for question answering over knowledge
  graphs.
\newblock In \emph{WWW}, pages 1191--1200.

\bibitem[{Bao et~al.(2016)Bao, Duan, Yan, Zhou, and
  Zhao}]{DBLP:conf/coling/BaoDYZZ16}
Jun{-}Wei Bao, Nan Duan, Zhao Yan, Ming Zhou, and Tiejun Zhao. 2016.
\newblock Constraint-based question answering with knowledge graph.
\newblock In \emph{COLING}, pages 2503--2514.

\bibitem[{Bast and Haussmann(2015)}]{DBLP:conf/cikm/BastH15}
Hannah Bast and Elmar Haussmann. 2015.
\newblock More accurate question answering on freebase.
\newblock In \emph{CIKM}, pages 1431--1440.

\bibitem[{Berant et~al.(2013)Berant, Chou, Frostig, and Liang}]{berant:13}
Jonathan Berant, Andrew Chou, Roy Frostig, and Percy Liang. 2013.
\newblock Semantic parsing on {Freebase} from question-answer pairs.
\newblock In \emph{EMNLP}, pages 1533--1544.

\bibitem[{Berant and Liang(2014)}]{DBLP:conf/acl/BerantL14}
Jonathan Berant and Percy Liang. 2014.
\newblock Semantic parsing via paraphrasing.
\newblock In \emph{ACL}, pages 1415--1425.

\bibitem[{Berant and Liang(2015)}]{berant2015agenda}
Jonathan Berant and Percy Liang. 2015.
\newblock Imitation learning of agenda-based semantic parsers.
\newblock \emph{TACL}, 3:545--558.

\bibitem[{Bordes et~al.(2015)Bordes, Usunier, Chopra, and Weston}]{bordes:15}
Antoine Bordes, Nicolas Usunier, Sumit Chopra, and Jason Weston. 2015.
\newblock Large-scale simple question answering with memory networks.
\newblock \emph{arXiv}.

\bibitem[{Cai and Yates(2013)}]{cai:13}
Qingqing Cai and Alexander Yates. 2013.
\newblock Large-scale semantic parsing via schema matching and lexicon
  extension.
\newblock In \emph{ACL}, pages 423--433.

\bibitem[{Chang and Manning(2012)}]{DBLP:conf/lrec/ChangM12}
Angel~X. Chang and Christopher~D. Manning. 2012.
\newblock Sutime: {A} library for recognizing and normalizing time expressions.
\newblock In \emph{LREC}, pages 3735--3740.

\bibitem[{Cui et~al.(2005)Cui, Sun, Li, Kan, and
  Chua}]{DBLP:conf/sigir/CuiSLKC05}
Hang Cui, Renxu Sun, Keya Li, Min{-}Yen Kan, and Tat{-}Seng Chua. 2005.
\newblock Question answering passage retrieval using dependency relations.
\newblock In \emph{{SIGIR}}, pages 400--407.

\bibitem[{Dietz and Gamari(2017)}]{dietz2017trec}
Laura Dietz and Ben Gamari. 2017.
\newblock {TREC CAR}: A data set for complex answer retrieval. version
  1.4.(2017).

\bibitem[{Dong et~al.(2017)Dong, Mallinson, Reddy, and
  Lapata}]{DBLP:conf/emnlp/DongMRL17}
Li~Dong, Jonathan Mallinson, Siva Reddy, and Mirella Lapata. 2017.
\newblock Learning to paraphrase for question answering.
\newblock In \emph{{EMNLP}}, pages 875--886.

\bibitem[{Fader et~al.(2014)Fader, Zettlemoyer, and
  Etzioni}]{DBLP:conf/kdd/FaderZE14}
Anthony Fader, Luke Zettlemoyer, and Oren Etzioni. 2014.
\newblock Open question answering over curated and extracted knowledge bases.
\newblock In \emph{KDD}, pages 1156--1165.

\bibitem[{Fader et~al.(2013)Fader, Zettlemoyer, and
  Etzioni}]{DBLP:conf/acl/FaderZE13}
Anthony Fader, Luke~S. Zettlemoyer, and Oren Etzioni. 2013.
\newblock Paraphrase-driven learning for open question answering.
\newblock In \emph{ACL}, pages 1608--1618.

\bibitem[{Ferrucci(2012)}]{ferrucci:12}
David~A. Ferrucci. 2012.
\newblock This is watson.
\newblock \emph{{IBM} Journal of Research and Development}, 56(3):1.

\bibitem[{Finkel et~al.(2005)Finkel, Grenager, and
  Manning}]{DBLP:conf/acl/FinkelGM05}
Jenny~Rose Finkel, Trond Grenager, and Christopher~D. Manning. 2005.
\newblock Incorporating non-local information into information extraction
  systems by gibbs sampling.
\newblock In \emph{ACL}, pages 363--370.

\bibitem[{Harabagiu et~al.(2006)Harabagiu, Lacatusu, and
  Hickl}]{DBLP:conf/sigir/HarabagiuLH06}
Sanda~M. Harabagiu, V.~Finley Lacatusu, and Andrew Hickl. 2006.
\newblock Answering complex questions with random walk models.
\newblock In \emph{{SIGIR}}, pages 220--227.

\bibitem[{Harabagiu et~al.(2003)Harabagiu, Maiorano, and
  Pasca}]{DBLP:journals/nle/HarabagiuMP03}
Sanda~M. Harabagiu, Steven~J. Maiorano, and Marius Pasca. 2003.
\newblock Open-domain textual question answering techniques.
\newblock \emph{Natural Language Engineering}, 9(3):231--267.

\bibitem[{Harabagiu et~al.(2001)Harabagiu, Moldovan, Pasca, Mihalcea, Surdeanu,
  Bunescu, Girju, Rus, and Morarescu}]{DBLP:conf/acl/HarabagiuMPMSBGRM01}
Sanda~M. Harabagiu, Dan~I. Moldovan, Marius Pasca, Rada Mihalcea, Mihai
  Surdeanu, Razvan~C. Bunescu, Roxana Girju, Vasile Rus, and Paul Morarescu.
  2001.
\newblock The role of lexico-semantic feedback in open-domain textual
  question-answering.
\newblock In \emph{ACL}, pages 274--281.

\bibitem[{Herrera et~al.(2004)Herrera, Pe{\~{n}}as, and
  Verdejo}]{DBLP:conf/clef/HerreraPV05}
Jes{\'{u}}s Herrera, Anselmo Pe{\~{n}}as, and Felisa Verdejo. 2004.
\newblock Question answering pilot task at {CLEF} 2004.
\newblock In \emph{{CLEF}}, pages 581--590.

\bibitem[{Joshi et~al.(2014)Joshi, Sawant, and
  Chakrabarti}]{DBLP:conf/emnlp/JoshiSC14}
Mandar Joshi, Uma Sawant, and Soumen Chakrabarti. 2014.
\newblock Knowledge graph and corpus driven segmentation and answer inference
  for telegraphic entity-seeking queries.
\newblock In \emph{EMNLP}, pages 1104--1114.

\bibitem[{Kocisk{\'{y}} et~al.(2017)Kocisk{\'{y}}, Schwarz, Blunsom, Dyer,
  Hermann, Melis, and Grefenstette}]{DBLP:journals/corr/abs-1712-07040}
Tom{\'{a}}s Kocisk{\'{y}}, Jonathan Schwarz, Phil Blunsom, Chris Dyer,
  Karl~Moritz Hermann, G{\'{a}}bor Melis, and Edward Grefenstette. 2017.
\newblock The narrativeqa reading comprehension challenge.
\newblock \emph{CoRR}, abs/1712.07040.

\bibitem[{Lai et~al.(2017)Lai, Xie, Liu, Yang, and
  Hovy}]{DBLP:conf/emnlp/LaiXLYH17}
Guokun Lai, Qizhe Xie, Hanxiao Liu, Yiming Yang, and Eduard~H. Hovy. 2017.
\newblock {RACE:} large-scale reading comprehension dataset from examinations.
\newblock In \emph{EMNLP}, pages 785--794.

\bibitem[{Magnini et~al.(2004)Magnini, Vallin, Ayache, Erbach, Pe{\~{n}}as,
  de~Rijke, Rocha, Simov, and Sutcliffe}]{DBLP:conf/clef/MagniniVAEPRRSS05}
Bernardo Magnini, Alessandro Vallin, Christelle Ayache, Gregor Erbach, Anselmo
  Pe{\~{n}}as, Maarten de~Rijke, Paulo Rocha, Kiril~Ivanov Simov, and Richard
  F.~E. Sutcliffe. 2004.
\newblock Overview of the {CLEF} 2004 multilingual question answering track.
\newblock In \emph{{CLEF}}, pages 371--391.

\bibitem[{Mohammed et~al.(2018)Mohammed, Shi, and
  Lin}]{DBLP:conf/naacl/MohammedSL18}
Salman Mohammed, Peng Shi, and Jimmy Lin. 2018.
\newblock Strong baselines for simple question answering over knowledge graphs
  with and without neural networks.
\newblock In \emph{NAACL-HLT}, pages 291--296.

\bibitem[{Prager et~al.(2004)Prager, Chu{-}Carroll, and
  Czuba}]{DBLP:conf/acl/PragerCC04}
John~M. Prager, Jennifer Chu{-}Carroll, and Krzysztof Czuba. 2004.
\newblock Question answering using constraint satisfaction:
  Qa-by-dossier-with-contraints.
\newblock In \emph{ACL}, pages 574--581.

\bibitem[{Rajpurkar et~al.(2018)Rajpurkar, Jia, and
  Liang}]{DBLP:conf/acl/RajpurkarJL18}
Pranav Rajpurkar, Robin Jia, and Percy Liang. 2018.
\newblock Know what you don't know: Unanswerable questions for squad.
\newblock In \emph{{ACL}}, pages 784--789.

\bibitem[{Rajpurkar et~al.(2016)Rajpurkar, Zhang, Lopyrev, and
  Liang}]{DBLP:conf/emnlp/RajpurkarZLL16}
Pranav Rajpurkar, Jian Zhang, Konstantin Lopyrev, and Percy Liang. 2016.
\newblock Squad: 100, 000+ questions for machine comprehension of text.
\newblock In \emph{EMNLP}, pages 2383--2392.

\bibitem[{Ravichandran and Hovy(2002)}]{DBLP:conf/acl/RavichandranH02}
Deepak Ravichandran and Eduard~H. Hovy. 2002.
\newblock Learning surface text patterns for a question answering system.
\newblock In \emph{ACL}, pages 41--47.

\bibitem[{Reddy et~al.(2014)Reddy, Lapata, and Steedman}]{reddy:14}
Siva Reddy, Mirella Lapata, and Mark Steedman. 2014.
\newblock Large-scale semantic parsing without question-answer pairs.
\newblock \emph{TACL}, pages 377--392.

\bibitem[{Saquete et~al.(2009)Saquete, Gonz{\'{a}}lez, Mart{\'{\i}}nez{-}Barco,
  Mu{\~{n}}oz, and Llorens}]{DBLP:journals/jair/SaqueteGMML09}
Estela Saquete, Jos{\'{e}} Luis~Vicedo Gonz{\'{a}}lez, Patricio
  Mart{\'{\i}}nez{-}Barco, Rafael Mu{\~{n}}oz, and Hector Llorens. 2009.
\newblock Enhancing {QA} systems with complex temporal question processing
  capabilities.
\newblock \emph{J. Artif. Intell. Res.}

\bibitem[{Saquete et~al.(2004)Saquete, Mart{\'{\i}}nez{-}Barco, Mu{\~{n}}oz,
  and Gonz{\'{a}}lez}]{DBLP:conf/acl/SaqueteMMG04}
Estela Saquete, Patricio Mart{\'{\i}}nez{-}Barco, Rafael Mu{\~{n}}oz, and
  Jos{\'{e}} Luis~Vicedo Gonz{\'{a}}lez. 2004.
\newblock Splitting complex temporal questions for question answering systems.
\newblock In \emph{ACL}, pages 566--573.

\bibitem[{Savenkov and Agichtein(2016)}]{savenkov:16}
Denis Savenkov and Eugene Agichtein. 2016.
\newblock {When a Knowledge Base Is Not Enough: Question Answering over
  Knowledge Bases with External Text Data}.
\newblock In \emph{SIGIR}, pages 235--244.

\bibitem[{Su et~al.(2016)Su, Sun, Sadler, Srivatsa, Gur, Yan, and
  Yan}]{DBLP:conf/emnlp/SuSSSGYY16}
Yu~Su, Huan Sun, Brian Sadler, Mudhakar Srivatsa, Izzeddin Gur, Zenghui Yan,
  and Xifeng Yan. 2016.
\newblock On generating characteristic-rich question sets for {QA} evaluation.
\newblock In \emph{EMNLP}, pages 562--572.

\bibitem[{Sun et~al.(2018)Sun, Dhingra, Zaheer, Mazaitis, Salakhutdinov, and
  Cohen}]{sun2018open}
Haitian Sun, Bhuwan Dhingra, Manzil Zaheer, Kathryn Mazaitis, Ruslan
  Salakhutdinov, and William~W Cohen. 2018.
\newblock Open domain question answering using early fusion of knowledge bases
  and text.
\newblock \emph{EMNLP}.

\bibitem[{Talmor and Berant(2018)}]{DBLP:journals/corr/abs-1803-06643}
Alon Talmor and Jonathan Berant. 2018.
\newblock The web as a knowledge-base for answering complex questions.
\newblock In \emph{NAACL}, pages 641--651.

\bibitem[{Trischler et~al.(2017)Trischler, Wang, Yuan, Harris, Sordoni,
  Bachman, and Suleman}]{DBLP:conf/rep4nlp/TrischlerWYHSBS17}
Adam Trischler, Tong Wang, Xingdi Yuan, Justin Harris, Alessandro Sordoni,
  Philip Bachman, and Kaheer Suleman. 2017.
\newblock Newsqa: {A} machine comprehension dataset.
\newblock In \emph{Rep4NLP@ACL}, pages 191--200.

\bibitem[{Trivedi et~al.(2017)Trivedi, Maheshwari, Dubey, and
  Lehmann}]{DBLP:conf/semweb/TrivediMDL17}
Priyansh Trivedi, Gaurav Maheshwari, Mohnish Dubey, and Jens Lehmann. 2017.
\newblock Lc-quad: {A} corpus for complex question answering over knowledge
  graphs.
\newblock In \emph{ISWC}, pages 210--218.

\bibitem[{Unger et~al.(2012)Unger, B{\"{u}}hmann, Lehmann, Ngomo, Gerber, and
  Cimiano}]{DBLP:conf/www/UngerBLNGC12}
Christina Unger, Lorenz B{\"{u}}hmann, Jens Lehmann, Axel{-}Cyrille~Ngonga
  Ngomo, Daniel Gerber, and Philipp Cimiano. 2012.
\newblock Template-based question answering over {RDF} data.
\newblock In \emph{WWW}, pages 639--648.

\bibitem[{Usbeck et~al.(2017)Usbeck, Ngomo, Haarmann, Krithara, R{\"o}der, and
  Napolitano}]{usbeck:17}
Ricardo Usbeck, Axel-Cyrille~Ngonga Ngomo, Bastian Haarmann, Anastasia
  Krithara, Michael R{\"o}der, and Giulio Napolitano. 2017.
\newblock {7th Open Challenge on Question Answering over Linked Data (QALD-7)}.
\newblock In \emph{SemWebEval}.

\bibitem[{Voorhees and Tice(2000)}]{DBLP:conf/sigir/VoorheesT00}
Ellen~M. Voorhees and Dawn~M. Tice. 2000.
\newblock Building a question answering test collection.
\newblock In \emph{{SIGIR}}, pages 200--207.

\bibitem[{Wang and Nyberg(2015)}]{DBLP:conf/acl/WangN15}
Di~Wang and Eric Nyberg. 2015.
\newblock A long short-term memory model for answer sentence selection in
  question answering.
\newblock In \emph{ACL}, pages 707--712.

\bibitem[{Xu et~al.(2016)Xu, Reddy, Feng, Huang, and
  Zhao}]{DBLP:conf/acl/XuRFHZ16}
Kun Xu, Siva Reddy, Yansong Feng, Songfang Huang, and Dongyan Zhao. 2016.
\newblock Question answering on {Freebase} via relation extraction and textual
  evidence.
\newblock In \emph{ACL}.

\bibitem[{Yahya et~al.(2013)Yahya, Berberich, Elbassuoni, and
  Weikum}]{DBLP:conf/cikm/YahyaBEW13}
Mohamed Yahya, Klaus Berberich, Shady Elbassuoni, and Gerhard Weikum. 2013.
\newblock Robust question answering over the web of linked data.
\newblock In \emph{CIKM}, pages 1107--1116.

\bibitem[{Yang et~al.(2014)Yang, Duan, Zhou, and
  Rim}]{DBLP:conf/emnlp/YangDZR14}
Min{-}Chul Yang, Nan Duan, Ming Zhou, and Hae{-}Chang Rim. 2014.
\newblock Joint relational embeddings for knowledge-based question answering.
\newblock In \emph{EMNLP}, pages 645--650.

\bibitem[{Yang et~al.(2015)Yang, Yih, and Meek}]{DBLP:conf/emnlp/YangYM15}
Yi~Yang, Wen{-}tau Yih, and Christopher Meek. 2015.
\newblock Wikiqa: {A} challenge dataset for open-domain question answering.
\newblock In \emph{EMNLP}, pages 2013--2018.

\bibitem[{Yao and Durme(2014)}]{DBLP:conf/acl/YaoD14}
Xuchen Yao and Benjamin~Van Durme. 2014.
\newblock {Information Extraction over Structured Data: Question Answering with
  Freebase}.
\newblock In \emph{ACL}, pages 956--966.

\bibitem[{Yin et~al.(2015)Yin, Duan, Kao, Bao, and Zhou}]{yin:15}
Pengcheng Yin, Nan Duan, Ben Kao, Junwei Bao, and Ming Zhou. 2015.
\newblock Answering questions with complex semantic constraints on open
  knowledge bases.
\newblock In \emph{CIKM}, pages 1301--1310.

\bibitem[{Ziegler et~al.(2017)Ziegler, Abujabal, Saha~Roy, and
  Weikum}]{ziegler:17}
David Ziegler, Abdalghani Abujabal, Rishiraj Saha~Roy, and Gerhard Weikum.
  2017.
\newblock Efficiency-aware answering of compositional questions using answer
  type prediction.
\newblock In \emph{IJCNLP}, pages 222--227.

\end{thebibliography}

\end{document}